\documentclass[letterpaper, 10 pt, conference]{ieeeconf}  

\IEEEoverridecommandlockouts                              

\usepackage{amsmath}
\usepackage{amssymb}
\usepackage{mathtools}

\newtheorem{theorem}{Theorem}[section]

\newtheorem{lemma}[theorem]{Lemma}
\newtheorem{corollary}[theorem]{Corollary}

\newtheorem{assumption}[theorem]{Assumption}

\newtheorem{remark}[theorem]{Remark}
\usepackage{subcaption}
\usepackage{float}
\usepackage{xcolor}

\allowdisplaybreaks

\newcommand{\norm}[1]{\left\lVert#1\right\rVert}

\newcommand{\y}{y}
\newcommand{\ysh}{y_{\Sigma}}

\newcommand{\p}{p}

\newcommand{\uu}{u}
\newcommand{\w}{w}
\newcommand{\e}{e}

\newcommand{\bt}{\boldsymbol{\tau}}

\title{\LARGE \bf
A finite-sample generalization bound for stable LPV systems}

\author{Dániel Rácz$^{1}$, Martin Gonzalez$^{2}$,
 Mihály Petreczky$^{3}$, András Benczúr$^{4}$ and Bálint Daróczy$^{4}$
\thanks{This research was supported by the European Union project
RRF-2.3.1-21-2022-00004 within the framework of the Artificial Intelligence
National Laboratory, by the C.N.R.S. E.A.I.  project "StabLearnDyn" and by
the E.D.F. project 101103386 "FaRADAI".
}
\thanks{$^{1}$HUN-REN SZTAKI and ELTE, Budapest, Hungary
        {\tt\small racz.daniel@sztaki.hun-ren.hu}}%
\thanks{$^{2}$Institut de Recherche Technologique SystemX, Palaiseau, France}
\thanks{$^{3}$
Univ. Lille, CNRS, Centrale Lille, UMR 9189 CRIStAL, F-59000 Lille, France}
\thanks{$^{4}$HUN-REN SZTAKI, Budapest, Hungary}
}

\begin{document}

\maketitle
\thispagestyle{empty}
\pagestyle{empty}

\begin{abstract}
One of the main theoretical challenges in learning dynamical
systems from data is providing upper bounds on  the generalization
error, that is, the difference between the expected prediction error and the
empirical prediction error measured on some finite sample. In machine learning,
a popular class of such bounds are the  so-called Probably Approximately Correct
(PAC) bounds.
In this paper, we derive a PAC bound for
stable continuous-time linear parameter-varying (LPV) systems. Our bound depends on the $H_2$ norm of the chosen class of the LPV systems, but does
not depend on the time interval for which the signals are considered.
\end{abstract}

\section{Introduction}
\label{sec:intro}

LPV \cite{toth2010modeling} systems are a popular class of
dynamical systems  control and system identification,
\textcolor{blue}{
e.g.
(\cite{Giarre2002,Verdult02,pigalpv,Lal11,toth2010modeling,OOMEN20121975,LopesDosSantos2008,CoxLPVSS}).}
Generally, LPV
systems are linear in state, input
and output signals, but the coefficients of these linear relationships depend on
the \emph{scheduling variables}.  These systems are popular due to their ability to model highly
non-linear phenomena while allowing much simpler theoretical
analysis. In this work, we consider \emph{continuous-time} LPV
systems in state-space form, where the system matrices are affine functions of
the scheduling variables. 

\textbf{Contribution.}
In learning and parametric system identification
our starting point is a
parametrized family of LPV systems, 
\textcolor{blue}{and our goal  is to find a member of this family, which predicts as accurately as possible the true output for each input and scheduling variable.}
\textcolor{blue}{To this end, }we assume that we have a dataset consisting
triplets of signals representing the input, the scheduling signal and the
corresponding  output of the true system, all defined on the finite time interval $[0,T]$, and measured during the learning/identification experiment. 
\textcolor{blue}{Learning/system identification algorithms then choose
a concrete element of the
parameterized family of LPV systems, which predicts the best outputs from this dataset used for learning. }
\begin{color}{blue}
However, for the identified
model to be useful, especially for control,
it should accurately predict the true output
for inputs and scheduling signals which were
not used during identification, i.e., it should \emph{generalize} well.
A widespread measure of generalization is the \emph{true loss}, i.e. the expected value of the prediction error, if the inputs and scheduling signals are sampled from the same distribution as in the dataset used for learning.  Unfortunately, the true loss is unknown in general, 
but it can be approximated by the \emph{empirical loss}, i.e. the
average of prediction errors by a certain model, where the average is taken over the elements of the data set. 

It is well-known that the empirical loss converges to the true loss as 
the number of data points grow, e.g., \cite[Lemma 8.2]{ljung1998system}.
However, for the purposes of evaluating 
the true loss based on the empirical loss, 
it is of interest to have a uniform bound, with respect to the parameterized family, on the difference between  the true loss and the empirical loss.  
Such uniform bounds are referred to as PAC bounds in the
literature \cite{alquier2021user, shalev2014understanding},
and they are a standard tool for theoretical analysis of statistical learning algorithms.  
\end{color}

In this paper, we derive  \textcolor{blue}{a particular PAC} bound for LPV systems.  
\begin{color}{blue}
The derived bound is of the magnitude $O(c/\sqrt{N})$, where
$c$ is a constant that depends on  the parametrization, and
$N$ is
the cardinality of an arbitrary finite sample used to evaluate or \textcolor{blue}{identify the model.}
The constant $c$ is 
the maximal $H_2$ norm of the elements of the parametrization.
In contrast to related results, the constant $c$, and thus the bound, does not depend on the length of the time interval $[0,T]$. 


\textbf{Using PAC bounds for learning.}
PAC bounds can be used in several ways \cite{alquier2021user,shalev2014understanding}, without claiming completeness, we mention the following applications:
\textbf{(1)} to evaluate 
the performance
of identified models based on their performance on a dataset (the training or the validation set),
\textbf{(2)} to analyze the effect of various quantities appearing in the bound (norms, VC-dimension, number of data points) on the generalization capability of the learned models, in particular, characterizing the amount of data
necessary for learning adequate models, 
\textbf{(3)} to select parametrizations which minimize the PAC bound and hence are more likely to allow learning models which generalize well and avoid overfitting.

\textbf{Motivation: system identification.}
PAC bounds are expected to have the same
applications for system identification as for learning, i.e. they will enable a sharper theoretical analysis
of system identification algorithms, by making explicit relationship between the empirical and true losses 
as functions of the number of data points and various properties of the parametrization. 
\end{color}

\textbf{Motivation: Machine Learning.}
\textcolor{blue}{There is consensus in the machine learning community on the usefulness of PAC bounds. Continuous-time
dynamical systems, e.g.\ continuous-time Recurrent Neural Networks (RNNs,
\cite{orvieto2023resurrecting}),  Neural Controlled Ordinary Differential
Equations (NCDEs,
\cite{kidger2022neural}) or structured
State-Space Models (SSMs, 
\cite{gu2023mamba})
are becoming increasingly popular
in machine learning. }
Since LPV systems include bilinear systems
\cite{isidori1985nonlinear} as a special case, in principle they could be used
as universal approximators for sufficiently smooth
dynamical systems \cite{KrenerApprox}, including important subclasses of RNNs
and NCDEs. Finally, LPV systems include linear state-space models, which are crucial ingredients of SSMs, and subclasses of
RNNs, hence PAC bounds for LPV systems are expected to be useful for PAC
bounds for NCDEs, RNNs and SSMs.

 \begin{color}{blue}
\textbf{Related work.}
PAC bounds for discrete-time linear systems were explored in system
identification in \cite{vidyasagar2006learning,campi2002finite}, but not for LPV
systems in state-space form and continuous-time. There are several PAC bounds
available for continuous-time RNNs and nonlinear systems
\cite{hanson2021learning,KOIRAN199863,kuusela2004learning,fermanian2021framing,marion2023generalization,hanson2024rademacher},
but these are exponential in the integration time.
The literature on finite-sample bounds for learning
discrete-time dynamical systems, e.g.
\cite{vidyasagar2006learning,simchowitz2019learning,oymak2021revisiting,Ziemann1}, considers different learning problems.
\end{color}

\textbf{Significance and novelty.} 
The main novelty 
is that 
\begin{color}{blue}
\textbf{(1)}
our error bound does not
depend on the length of the integration interval $T$,
\textbf{(2)} it exploits quadratic stability,
\textbf{(3)} it uses $H_2$ norm  defined via Volterra-kernels to estimate the Rademacher complexity.
This is in contrast \cite{hanson2024rademacher,hanson2021learning,fermanian2021framing,marion2023generalization} which used Fliess-series expansions for that purpose. 
It was precisely the use of  Volterra-series and $H_2$ norms which allowed us to formulate bounds which do not grow with $T$. In  turn, the latter
is important as dynamical
systems are often used for making long-term predictions.
\end{color}

\textcolor{blue}{\textbf{Structure of the paper.} Next, in Section~\ref{sec:lpv} we set notations and definitions related to the LPV system, then in Section~\ref{sec:learning} we define the problem of generalization in case of LPV systems, when we can only measure the error of the system on a finite set of sequences. In our case this set is a particular set of LPV systems, which meet the stability and other crucial conditions for having a time-invariant bound. We define these conditions and the $H_2$ norm of output in Section~\ref{sec:ass}. We state our time-independent bound in Section~\ref{sec:main} for the generalization gap, the difference between the true error and the error measured on the finite set of sequences, based on the Rademacher complexity of the hypothesis set.  We prove our bound in Section~\ref{sec:proofsketch} by showing that under our assumptions the LPV system has finite $H_2$ norm, thus we may upper bound the Rademacher complexity of the system. Finally, in Section~\ref{sec:experiments} we consider a system and a dataset,  we estimate the elements of the bound and show that the bound in this case is meaningful. } 

\section{LPV systems}
\label{sec:lpv}

In this paper, we consider \emph{LPV state-space representations
with affine dependence on the parameters (LPV-SSA)}, i.e. systems of the form
\begin{equation}
    \label{system:2}
     \Sigma 
    \begin{cases}
      \dot{x}(t) = A(p(t)) x(t) + B(p(t))u(t), ~  x(0)=0 \\
      y_{\Sigma}(t)=C(p(t)) x(t) \\
    \end{cases}
\end{equation}
where $x(t) \in \mathbb{R}^{n_{\text{x}}}$ is the state vector, $u(t) \in
 \mathbb{R}^{n_{\text{in}}}$ is the input and 
 $y_{\Sigma}(t) \in \mathbb{R}^{n_{\text{out}}}$ is the output of the system for all $t \in [0, T]$, and $n_{\text{x}}, n_{\text{in}}, n_{\text{out}} \in \mathbb{N}^{+}$. The
 vector $p(t) = (p_1(t), \dots, p_{n_{\text{p}}}(t))^T \in \mathbb{P} \subseteq
 \mathbb{R}^{n_{\text{p}}}$ is the scheduling variable for $n_{\text{p}} \in \mathbb{N}^{+}$. \textcolor{blue}{Note that $n_p$ is the number of scheduling variables in the
system, therefore it is the most important parameter regarding
the complexity of the system.} The matrices 
 of the system $\Sigma$
 are assumed to depend on $p(t)$ affinely, i.e.
 $A(p(t)) = A_0 + \sum\limits_{i=1}^{n_{\text{p}}}p_i(t)A_i$,
 $B(p(t)) = B_0 + \sum\limits_{i=1}^{n_{\text{p}}}p_i(t)B_i$ and
 $C(p(t)) = C_0 + \sum\limits_{i=1}^{n_{\text{p}}}p_i(t)C_i$
 for matrices $A_i \in \mathbb{R}^{n_{\text{x}}
 \times n_{\text{x}}}, B_i \in \mathbb{R}^{n_{\text{x}} \times n_{\text{in}}}$ and  $C_i \in
 \mathbb{R}^{n_{\text{out}} \times n_{\text{x}}}$, $i=0,\ldots,n_{\text{p}}$, which do not depend on time
 or the scheduling signal. 
 We identify the LPV-SSA 
 $\Sigma$ with the tuple $\Sigma=(A_i,B_i,C_i)_{i=0}^{n_{\text{p}}}$.

A \emph{solution} of $\Sigma$ refers to the tuple 
of functions $(u, p, x, y_{\Sigma})$,
all defined on 
$[0,T]$, such that
$x$ is absolutely continuous, 
$y_{\Sigma}$, $u$ and $p$ are
piecewise continuous  and they satisfy \eqref{system:2}.
Note that the output $y_{\Sigma}$ is uniquely
determined by $u$ and $p$,
since the initial state $x(0)$ is set to zero. To emphasize this dependence, 
we denote $y_{\Sigma}$ by $y_{\Sigma}(u,p)$. %
Thus, the  scheduling signal $p(t)$ 
behaves as an external input, too.

For the sake of compactness we make a series of
simplifications. First, as already stated, the initial state is set to zero.
This is not a real restriction, as we consider stable systems for which the
contribution of the non-zero initial state decays exponentially. Second, we
work with systems with scalar output, i.e. let $n_{\text{out}} = 1$.  Third, we assume
that the scheduling variables take values in $\mathbb{P} \subseteq
[-1,1]^{n_{\text{p}}}$. This is a standard assumption in the literature
\cite{toth2010modeling} and it can always be achieved by an affine
transformation, if the scheduling variables take values in a suitable interval.

\section{Learning problem and generalization}
\label{sec:learning}

We now define the learning problem
for LPV systems along the lines of classical statistical learning
theory \cite{shalev2014understanding}.
For this purpose, let us fix a time interval $[0,T]$ and a set 
$\mathcal{E}$ of LPV systems of the form \eqref{system:2}.

Let $\mathcal{U}$, $\mathcal{P}$ and $\mathcal{Y}$ be sets of piecewise
continuous functions defined on $[0, T]$ and taking values in
$\mathbb{R}^{n_{\text{in}}}$, $\mathbb{P}$ and $\mathbb{R}^{n_{\text{out}}}$ respectively.
Hereinafter we use the standard terminology of probability
theory \cite{Bilingsley}. Consider the probability space $(\mathcal{U} \times
\mathcal{P} \times \mathcal{Y}, \mathcal{B},\mathcal{D})$,  where $\mathcal{B}$
is a suitable $\sigma$-algebra and $\mathcal{D}$ is a probability measure on
$\mathcal{B}$. For example, $\mathcal{B}$ could be the direct product of the
standard cylindrical Borel $\sigma$-algebras defined on the function spaces
$\mathcal{U},\mathcal{P}$ and $\mathcal{Y}$.
Let us denote by $\mathcal{D}^N$ the $N$-fold product measure of $\mathcal{D}$
with itself. We use $E_{(u,p,y) \sim \mathcal{D}}$, $P_{(u,p,y) \sim
\mathcal{D}}$, $E_{S \sim \mathcal{D}^N}$ and $P_{S \sim
\mathcal{D}^N}$ to denote expectations and probabilities w.r.t. the measures
$\mathcal{D}$ and $\mathcal{D}^N$ respectively. The notation $S \sim
\mathcal{D}^N$ tacitly assumes that $S \in (\mathcal{U}
\times \mathcal{P} \times \mathcal{Y})^N$, i.e. $S$ is made of $N$ triplets of
input, scheduling and output trajectories. Intuitively, we
think of $S$ as a dataset of size N drawn randomly and
independently from the distribution $\mathcal{D}$.

Consider a \emph{loss} function $\ell: \mathbb{R} \times \mathbb{R} \rightarrow
[0,+\infty)$, which measures the discrepancy between two possible output values.
Some of the widespread choices are $\ell(a,b)=\|a-b\|$ or
$\ell(a,b)=\|a-b\|^2$. 

The learning objective is to find an LPV system $\Sigma \in
\mathcal{E}$ such that the \textit{true risk at time $T$}, defined as 
\begin{align*}
  \mathcal{L}(\Sigma) =
  \mathbb{E}_{(u, p, y) \sim \mathcal{D}}[\ell(
  y_{\Sigma}(u,p)(T),y(T))]
\end{align*}
is as small as possible.  
Since the distribution $\mathcal{D}$ is unknown, minimizing the true risk is
impossible. Therefore, the true risk is approximated by the \textit{empirical
risk at time $T$} w.r.t. a dataset $S = \{(u_i, p_i, y_i) \}_{1 \leq i \leq N}$,
defined as
\begin{align*}
    \mathcal{L}_{N}^{S}(\Sigma) =
    \frac{1}{N}\sum\limits_{i = 1}^{N} \ell(\y_{\Sigma}(u_i, p_i)(T),y_i(T)). 
\end{align*}
In practice, selecting an appropriate model is done by minimizing the empirical
risk w.r.t. a so-called training dataset, while the trained model is usually
evaluated by computing the empirical risk w.r.t. a separate test dataset. In
both cases, we need a bound on the difference between the true and the empirical
risk.  
\textcolor{blue}{That is, we need to bound }
the \emph{generalization gap}, defined as $\sup_{\Sigma
\in \mathcal{E}}( \mathcal{L}(\Sigma) - \mathcal{L}_N^{S}(\Sigma))$.

\begin{remark}
In the data, the scheduling signal $\p$ may depend on
$\uu$ or $y$, and the data may be generated by a quasi-LPV system. However, for
the models to be learnt the scheduling signal acts as an input, \textcolor{blue}{as it is customary}
in system identification for LPV systems
\cite{toth2010modeling}.
\end{remark}

\section{Technical preliminaries and assumptions}
\label{sec:ass}
We start by presenting a Volterra-series representation of the output of an LPV
system of the form \eqref{system:2}, which plays a central role in
formulating and proving the main result. To this end, we introduce the following
notation.

\textbf{Notation (Iterated integrals).} 
Let $\Delta_k^{t} = \left\{(\tau_k,\dots,\tau_1) \mid t \ge \tau_k \geq \dots
\geq \tau_1 \geq 0 \right\} \subset \mathbb{R}^{k}$ for any positive integer $k$.
Moreover, let $\Delta_k^{\infty} = \left\{(\tau_k,\dots,\tau_1) \mid \tau_k \geq
\dots \geq \tau_1 \geq 0 \right\}$. Clearly, $\Delta_k^{t} \subseteq
\Delta_k^{\infty}$ for all $t \in [0,+\infty)$. For $t \in \mathbb{R}$ and $\tau
= (\tau_k,\ldots,\tau_1)$ we use the notation $(t, \tau) = (t, \tau_k, \ldots,
\tau_1)$. In addition, we use the following notation for
iterated integrals (for any function $f$ for which the integrals are well
defined), for $t \in [0,+\infty]$
\begin{align*}
    \int_{\Delta_{k}^{t}} f(\tau) \,d \tau \equiv 
    \int_{0}^{t} \int_{0}^{\tau_k} \cdots \int_{0}^{\tau_2}
    f(\tau_k,\ldots,\tau_1)
    \,d \tau_1 \cdots \,d \tau_k.
\end{align*}

Let $[n] = \{1,\ldots,n\}$ and $[n]_0 = \{0,\ldots,n\}$ for all $n \in
\mathbb{N}$. Let $I_k = [n_{\text{p}}]^k$ be the set of multi-indices of length $k$,
so an element $I$ of $I_k$ is a tuple of the form
$I=(i_1,\ldots,i_k)$. By slight abuse of notation,
let $I_0$ be the singleton set $\{\emptyset\}$. For a system $\Sigma$ of the
form \eqref{system:2}, for every $I \in I_k$, $t \in [0,+\infty)$,
$i_q,i_r \in [n_{\text{p}}]_0$, and for every $p \in \mathcal{P}$, $u \in \mathcal{U}$
and $\lambda \ge 0$, we define the
\emph{$\lambda$-weighted LPV Volterra-kernels} $w_{i_q,i_r,I}^{\lambda}:
\Delta_{k+1}^{\infty} \rightarrow \mathbb{R}^{1 \times n_{\text{in}}}$, the
\emph{scheduling product} $p_{I}: \Delta_{k+1}^T \rightarrow
\mathbb{R}$ and the \emph{$\lambda$-weighted scheduling-input product}
$\varphi_{i_q,i_r,I}^{\lambda}: \Delta_{k+1}^T \rightarrow
\mathbb{R}^{n_{\text{in}}}$ as follows. 
For $k=0$ and 
$I=\emptyset$, for any $\tau=(\tau_1) \in \Delta_1^{t}=[0,t]$, set
\begin{align*}
& w^{\lambda}_{i_q, i_r, \emptyset}(\bt)
:=C_{i_q}e^{A_0\tau_1}B_{i_r}e^{\frac{\lambda}{2}\tau_1}, ~
p_{\emptyset}(\tau):=1, \\
& \varphi_{i_q,i_r,\emptyset}^{\lambda}(\tau):=
p_{i_q}(T)p_{i_r}(T-\tau_1)u(T-\tau_1)e^{-\frac{\lambda}{2}\tau_1}.
\end{align*}
For $k > 0$, $\tau=(\tau_{k+1},\ldots,\tau_1) \in \Delta_{k+1}^{t}$, 
$t=\infty$ or $t=T$ respectively
and $I=(i_1,\ldots,i_k)$ set
 \begin{align*}
     & w_{i_q, i_r, I}^{\lambda}(\tau) :=  \\
     & C_{i_q} e^{A_0 (\tau_{k+1}-\tau_k)} A_{i_k}e^{A_0(\tau_k-\tau_{k-1})} 
     \cdots A_{i_1}e^{A_0 \tau_1}B_{i_r}e^{\frac{\lambda}{2} \tau_{k+1}},\\
     & p_{I}(\tau):=
      \prod_{j=1}^{k} p_{i_j}(\tau_j+T-\tau_{k+1}),  \\
     & \varphi^{\lambda}_{i_q,i_r,I}(\tau):=
      p_{i_q}(T)p_{i_r}(\tau_{k+1})\p_{I}(\tau)u(T-\tau_{k+1})
      e^{-\frac{\lambda}{2}\tau_{k+1}}.
 \end{align*}

 In the definition above, the value $k$ is always equal to the size of the tuple
 $I$. The respective domains of these functions each contain tuples of size $k+1$
 (denoted by $\tau$).
 \textcolor{blue}{The $\lambda$-weighted LPV Volterra kernels 
 represent 
 weighted Volterra kernels of certain
 bilinear systems, outputs of which 
 determine the output of \eqref{system:2}. The $\lambda$-weighted scheduling-input products capture the polynomial relationship between the outputs of these bilinear systems and the scheduling and input signals. The weighting was introduced in order to make the series of the $L^2$ norms of these products  square summable. 
The terms belonging to $i_r$ and $i_q$ are related to the
effect of $B_{i_r}$ and $C_{i_q}$ on the output of 
\eqref{system:2}.
 The following Lemma captures this intuition in a rigorous way.}

\begin{lemma}
\label{lemma:volterra}
    For every $(u,p) \in \mathcal{U} \times \mathcal{P}$, $\lambda \ge 0$, the
    output of $\Sigma$ at time $T$ admits the following representation:
    \begin{align*}
        &y_{\Sigma}(u, p)(T) \!\! = \!\!  
        \sum\limits_{i_q, i_r = 0}^{n_{\text{p}}}
        \sum\limits_{k=0}^{\infty}
        \sum\limits_{I \in I_k}
        \int_{\Delta_{k+1}^{T}}\!\!\!\!
        w^{\lambda}_{i_q, i_r, I}(\tau)\varphi^{\lambda}_{i_q,i_r,I}(\bt)
        d\tau.
    \end{align*}
\end{lemma}
The proof is based on a Volterra series expansion \cite{isidori1985nonlinear}
and can be found \textcolor{blue}{in \cite[Appendix A]{arxiv}.}
This observation allows us to
represent the output of an LPV system as a scalar product in a suitable Hilbert
space which turns out to be the key for the proof of the main result. 

Next, we define the \emph{$\lambda$-weighted $H_2$ norm}, a variant of the classical
$H_2$ norm, parameterized by a constant $\lambda > 0$,  
\begin{align*}
        \norm{\Sigma}_{\lambda,H_2}^2 \!\! :=&  \!\!
         \sum_{i_q,i_r=0}^{n_{\text{p}}} 
         \sum_{k=0}^{\infty}
         \sum_{I \in I_k}
         \int_{\Delta_{k+1}^{\infty}}
         \norm{w_{i_q, i_r, I}^{\lambda}(\tau)}_{2}^2 d\tau.
\end{align*}
\begin{color}{blue}
\begin{remark}
    The norm $\norm{\Sigma}_{\lambda,H_2}$ is finite, if the Volterra kernels $w_{i_q, i_r, I}^{\lambda}$ are finite energy signals and
the sum of their $L_2$ norms is finite too. For $\lambda=0$,
the definition $\norm{\Sigma}_{\lambda,H_2}$ is a direct extension for bilinear systems \cite{zl02}.
When applied to LTI systems with $\lambda=0$, the above defined norm is the $H_2$ norm. For LPV systems there are several possible definitions of $H_2$ norms, which are not equivalent, not even for linear, time-varying systems, for an overview see \cite[Section 2.2]{SZNAIER2002957}.  The norm $\norm{\Sigma}_{\lambda,H_2}$
is
different from these other $H_2$ norms for LPV systems. 
However, as  it is shown in Lemma \ref{lemma:h2norm} below, under certain stability conditions this norm exists, and  similarly to other $H_2$ norms, it is an upper bound on 
peak output under unit energy input.
We use this norm instead of other $H_2$ norms to succesfully upper bound the Rademacher complexity in the proof of the main result.
\end{remark}


In order to state the main result of the paper, we need
to state several assumptions.
To this end, we introduce the following notation.
\textcolor{blue}{Let us
denote by $L^2([0,T],\mathbb{R}^{n_{\text{in}}})$ the space of all measurable functions
$f:[0,T] \rightarrow \mathbb{R}^{n_{\text{in}}}$ such that
$\norm{f}^2_{L^2([0,T],\mathbb{R}^{n_{\text{in}}})}:=\int_0^T \|f(t)\|_2^2 dt$ is
finite. }

%
\begin{assumption}
\label{ass:tech}
 \textbf{Stability.}  
    There exists  $\lambda \geq n_{\text{p}}$ such that for any $\Sigma \in \mathcal{E}$
    of the form \eqref{system:2} there exists $Q \succ 0$   such that 
    \begin{align}   
    \label{ass:stability}    
        & A_0^T Q + Q A_0 + \sum\limits_{i=1}^{n_{\text{p}}} A_i^T Q A_i +     
        \sum\limits_{i=0}^{n_{\text{p}}} C_i^T C_i 
        \prec -\lambda Q.
    \end{align}
  \textbf{Bounded $H_2$ norm.}
    We assume that $\sup_{\Sigma \in \mathcal{E}} \norm{\Sigma}_{\lambda,H_2}
    \leq c_{\mathcal{E}}$. 
\\    
  \textbf{Bounded signals.}
    For any $u \in \mathcal{U}$ and $y \in
    \mathcal{Y}$, $\norm{u}_{L^2([0, T],\mathbb{R}^{n_{\text{in}}})} \leq L_{u}$
    and $|y(T)| \leq c_{y}$. 
 \\
 \textbf{Lipschitz loss function.}
    The loss function $\ell$ is $K_{\ell}$-Lipschitz-continuous, i.e.
    $|\ell(y_1,y_1')-\ell(y_2,y_2')| \le K_{\ell}(|y_1-y_2|+|y_1'-y_2'|)$
    for all $y_1,y_2,y_1',y_2' \in \mathbb{R}$, and
    $\ell(y,y)=0$ for all $y \in \mathbb{R}$.
\end{assumption}
\textbf{Discussion on the assumptions.}
The first assumption ensures quadratic stability of all LPV systems 
in the parametrization 
with a decay rate $\frac{\lambda}{2}$, and it ensures
the finiteness of $H_2$ norms:
\begin{lemma}
\label{lemma:h2norm}
     If 
     \eqref{ass:stability} 
     with $\lambda \ge n_{\text{p}}$ and $Q \succ 0$, then for any $\Sigma \in \mathcal{E}$, 
      \(   \norm{\Sigma}_{\lambda,H_2}^2 
         \le \sum_{i_r=0}^{n_{\text{p}}} \mathrm{trace} (B_{i_r}^T Q B_{i_r}) < +\infty. \)
     Additionally, for any $u \in \mathcal{U}$, $p \in \mathcal{P}$, we have
  \begin{center}
     $|y_{\Sigma}(u,p)(T) | \le (n_{\text{p}}+1)\norm{\Sigma}_{\lambda,H_2}
     \norm{u}_{L^2([0,T],\mathbb{R}^{n_{\text{in}}})}.$
  \end{center}
\end{lemma}
The proof can be found \textcolor{blue}{in \cite[Appendix B]{arxiv}.} 

The first assumption is not restrictive as it can be translated into inequalities on the eigenvalues of the system matrices \cite[Appendix B]{arxiv}, which are not too difficult to ensure by choosing a suitable parametrization.

The second assumption is that the $\lambda$-weighted $H_2$ norm of the elements of
class $\mathcal{E}$ are bounded by $c_{\mathcal{E}}$. This assumption holds for instance for continuous parametrizations with a bounded compact parameter set.

The third assumption means that the input signal has a finite energy and the  output signal is bounded. 
The assumption on $|y(T)|$ means the true labels are bounded. By Lemma \ref{lemma:h2norm}, the first two assumptions along with finite energy inputs already imply that the outputs of the 
systems from $\mathcal{E}$ are bounded uniformly by a suitable $c$.
In practice, considering finite energy inputs and bounded outputs is often natural. 

The last assumption is standard in machine learning, it is satisfied for $\ell(y,y^{'})=$$|y-y^{'}|$, and even
for the square loss, if the latter is restricted to bounded labels.

%

\end{color}

 %
%

\section{Main result}
\label{sec:main}

We are ready to state our main contribution.
\begin{theorem}[Main result]
\label{thm:main}
    Let $c := 2 K_{\ell} \max \{ L_{u}(n_{\text{p}} + 1)c_{\mathcal{E}}, c_{y} \}$
    and $R(\delta) := c \left (2 + 4
    \sqrt{2\log(4/\delta)}\right )$.
    Under Assumptions 
    \textcolor{blue}{ \ref{ass:tech}}
    , for any $\delta \in (0,1)$, we have
    \begin{align*}
        \mathbb{P}_{\mathbf{S} \sim
        \mathcal{D}^N}\Bigg(\forall \Sigma \in \mathcal{E}:
        \mathcal{L}(\Sigma) - \mathcal{L}^{S}_{N}(\Sigma)
        \leq \frac{R(\delta)}{\sqrt{N}} \Bigg) \geq 1 - \delta.
    \end{align*}
\end{theorem}




\begin{color}{blue}
\textbf{Application of PAC to system identification.}
Any system identification 
algorithm maps a dataset $S$ to a model
$\hat{\Sigma}=\hat{\Sigma}(S)$. As a PAC bound holds uniformly on
all models, with  probability at least $1-\delta$ over $S$,
we get the explicit high-probability bound on the true error 
\[ \mathcal{L}(\hat{\Sigma}) \le  \mathcal{L}_{N}^{S}(\hat{\Sigma}) +
\frac{R(\delta)}{\sqrt{N}}. \]
This allows us to evaluate the prediction error of the identified model for unseen data. Moreover,  for any accuracy $\epsilon > 0$ we can determine the minimum number of data points $N_m=\frac{(R(\delta))^2}{\epsilon^2}$ such that  if $N \ge N_m$, then the true loss of \emph{any} identified model is smaller than $\epsilon$ plus the empirical loss. 
The integer $N_m$ 
represents the minimal number of data points after which we can view the empirical loss as indicative of the true loss.
In fact, using \cite[Theorem 26.5]{shalev2014understanding}
and the upper bound on the Rademacher complexity of $\mathcal{E}$ from the proof of Theorem \ref{thm:main},
we can get a high-probability upper bounds on the difference
between the true loss of the minimal prediction error model $\hat{\Sigma}=\mathrm{argmin}_{\Sigma \in \mathcal{E}}\mathcal{L}^S_N(\Sigma)$  and the best possible model:  $\Sigma_{\star}=\mathrm{argmin}_{\Sigma \in \mathcal{E}} \mathcal{L}(\Sigma)$, see \cite[Appendix E]{arxiv}.


\end{color}
\begin{color}{blue}
\textbf{Comparison with system identification.}
Classical results in system identification aim at showing
that  for large enough $N$, an identification algorithm which choose models with a small empirical loss small 
will result in models with a small true loss, 
e.g. \cite[Lemma 8.2]{ljung1998system} for LTI systems.
However,  in contrast to Theorem \ref{thm:main} these classical results say little about how large $N$ should be so that the empirical loss upper bounds the true loss with a certain accuracy.

In contrast to system identification, we assume access to several i.i.d. samples 
of input, output and scheduling signals. 
While it is somewhat restrictive, it is still applicable in many scenarios.
Furthermore, deriving PAC bounds in this setting is a first step towards PAC bounds for the case of a single long time signal.

\textbf{Sampling, persistence of excitation, etc.}
Since PAC bounds hold for any identification algorithms, we did not make any assumptions on such, otherwise crucial issues, as persistence of excitation, sampling, etc. 


\textbf{Parameter estimation.}
For linear systems it is known that models with a small prediction error tend to be close to the true one, 
under suitable identifiability assumptions
\cite[Theorem 8.3]{ljung1998system}. 
 For LPV systems, this problem requires further research, but we expect similar results.
\end{color}

\textbf{Discussion on the bound: dependence on $N$ and $T$.}
The bound in Theorem \ref{thm:main} tends to zero as $N$ grows to infinity and
is also independent of the integration time $T$, a consequence of assuming
stability of the models. The latter is a significant improvement compared to
prior work \cite{fermanian2021framing,marion2023generalization,
hanson2021learning,hanson2024rademacher}. We conjecture that some form of
stability is also necessary for time-independent bounds, as intuitively in case
of unstable systems small modelling errors may lead to a significant increase of
the prediction error in the long run.
The bound grows linearly
with the maximal $H_2$ norm of the elements of $\mathcal{E}$,
with  the  maximal possible value of the true outputs, and  with 
the maximal energy of the inputs.
\begin{color}{blue}

\textbf{Multiple outputs.}
Under suitable assumptions on the loss function,
which are satisfied for quadratic and $1$-norm losses,
the results can be extended to multiple outputs by applying Theorem \ref{thm:main} 
separately for each output component,  
see \cite[Appendix C]{arxiv} for a detailed derivation.

\end{color}



\section{Proof of Theorem \ref{thm:main}}
\label{sec:proofsketch}
The main component of the proof is the estimation of the
Rademacher complexity of the class of LPV systems.

\textcolor{blue}{Recall from\cite[Def. 26.1]{shalev2014understanding}}  that Rademacher complexity of a bounded set $\mathcal{A} \subset
     \mathbb{R}^{m}$ 
    is defined as
    \begin{align*}
        R(\mathcal{A}) = 
        \mathbb{E}_{\sigma}\Bigg[\sup_{a \in \mathcal{A}}
        \frac{1}{m} \sum\limits_{i = 1}^{m}\sigma_i a_i \Bigg],
    \end{align*}
    where the random variables $\sigma_i$ are i.i.d such that
    $\mathbb{P}[\sigma_i = 1]  = \mathbb{P}[\sigma = -1] = 0.5$. The Rademacher
    complexity of a set of functions $\mathcal{F}$ over a set of samples $S =
    \{s_1\dots s_m\}$ is defined as $R_{S}(\mathcal{F}) =
    R(\left\{(f(s_1),\dots,f(s_m)) \mid f \in \mathcal{F} \right\}).$

Intuitively, Rademacher complexity measures the richness of a set of functions, see
e.g. chapter 26 in \cite{shalev2014understanding}, and can be used for deriving
PAC bounds \cite[Theorem 26.5]{shalev2014understanding} for general models.
Below we restate this result for LPV systems.
\begin{theorem}    \label{pac} 
    Let $L_0(T)$ denote the set of functions of the from
    $(u,p,y) \mapsto \ell(\ysh(u,p)(T),y(T))$ for $\Sigma
    \in \mathcal{E}$.
    Let $B(T)$ be such that 
    the functions from $L_0(T)$ all take values from the interval $[0,B(T)]$.
    Then  for any $\delta \in (0, 1)$  we have
    \begin{align*}
        \mathbb{P}_{S \sim \mathcal{D}^N}\Bigg(\forall \Sigma \in \mathcal{E}:
        \mathcal{L}(\Sigma) - \mathcal{L}^{S}_{N}(\Sigma)
         \leq R_{S}^{T, N, \delta} \Bigg) \geq 1 - \delta
    \end{align*}
    where $R_{S}^{T, N, \delta} = 2R_{S}(L_{0}(T)) $+
    4$B(T)\sqrt{\frac{2 \log (4/\delta)}{N}}$.
\end{theorem}

The proof of Theorem \ref{thm:main} follows from Theorem \ref{pac}, by first
bounding the Rademacher complexity of $R_{S}(L_0(T))$ and then bounding
the constant $B(T)$.

\textbf{Step 1: Showing $R_{S}(L_0(T)) \leq \frac{c}{\sqrt{N}}$.}
Consider the class $\mathcal{F}$ of output response functions $(u,p) \mapsto
\ysh(u,p)(T)$ for $\Sigma \in \mathcal{E}$ and the corresponding Rademacher
complexity $R_{\mathbf{S}}(\mathcal{F})$. By \cite[Lemma
26.9]{shalev2014understanding} and Assumption \textcolor{blue}{\ref{ass:tech}},
$R_{S}(L_0(T)) \le K_{\ell} R_{S}(\mathcal{F})$, hence it is
enough to bound $R_{S}(\mathcal{F})$. For the latter, we need the
following Lemma.

\begin{lemma}
\label{lemma:main} 
    There exists a Hilbert space $\mathcal{H}$ such that for every $\Sigma \in
    \mathcal{E}$ there exists  $\w^{T, \Sigma} \in \mathcal{H}$ and for every
    $(u,p) \in \mathcal{U} \times \mathcal{P}$ there exists
    $\varphi^{T,u,p} \in \mathcal{H}$, such that 
    $$y_{\Sigma}(u, p)(T) = \langle \w^{T, \Sigma},\varphi^{T, \uu,\p}
    \rangle_{\mathcal{H}}, $$ and $\norm{\varphi^{T, u, p}}_{\mathcal{H}}
    \leq L_{u} (n_{\text{p}} + 1)$ and $\norm{\w^{T, \Sigma}}_{\mathcal{H}} \leq
    \norm{\Sigma}_{\lambda, H_2}$.
\end{lemma}

\textcolor{blue}{
\begin{altproof}
\textbf{Proof.} 
Let $\mathcal{V}$ be the vector space
  consisting of sequences of the form $f=\{f_{i_q,i_r,I} \mid
  I \in I_k; i_q, i_r \in [n_{\text{p}}]_0\}_{k=0}^{\infty}$ such that
  $f_{i_q,i_r,I}:\Delta_{k+1}^{T} \mapsto \mathbb{R}^{1 \times n_{\text{in}}}$
  is measurable. For any $f,g \in \mathcal{V}$ let us define the series
  \begin{align*}
    &\langle f, g \rangle =
    \sum_{i_q,i_r=0}^{n_{\text{p}}}\sum_{k=0}^{\infty} \sum_{I \in I_k} 
    \int_{\Delta_{k+1}^T} f_{i_q,i_r,I}(\tau) g_{i_q,i_r,I}(\tau)^T d\tau.
  \end{align*}
  and for any $f \in \mathcal{V}$, let us denote by $\|f\|^2=\langle f,f
  \rangle$. Let $\mathcal{H}$ consists of those element $f \in \mathcal{V}$ for
  which the series $\|f\|^2$ is convergent. Then for any $f,g \in \mathcal{H}$,
  the series $\langle f,g \rangle$ is absolutely convergent. Let us denote its
  limit by $\langle f,g \rangle_{\mathcal{H}}$. Then $\mathcal{H}$ is a
  Hilbert-space with the scalar product $\langle \cdot, \cdot
  \rangle_{\mathcal{H}}$, and we denote by $\norm{\cdot}_{\mathcal{H}}$ the
  corresponding norm. 
  Let $\lambda \ge n_{\text{p}}$ be such that Assumption \ref{ass:tech} is satisfied. Let
  \begin{align*}
    &\w^{T,\Sigma} = 
    \{(\w_{i_q,i_r,I}^{\lambda})|_{\Delta_{k+1}^{T}}
    \mid I \in I_k; i_q, i_r \in [n_{\text{p}}]_0\}_{k=0}^{\infty}  \\ 
    & \varphi^{T,u, p}=\{(\varphi^{\lambda}_{i_q, i_r, I})^T 
    \mid I \in I_k; i_q, i_r \in [n_{\text{p}}]_0\}_{k=0}^{\infty}.
  \end{align*}
  We show that $\w^{T, \Sigma} \in \mathcal{H}$ and $\varphi^{T, \uu,
  \p} \in \mathcal{H}$ by proving that $\norm{\w^{T,
  \Sigma}}_{\mathcal{H}}^2$ and $\norm{\varphi^{T, \uu,
  \p}}_{\mathcal{H}}^2$ are finite and bounded as claimed in the Lemma.
  By the definition of $w^{T,\Sigma}$ it is clear that
  $\norm{\w^{T, \Sigma}}_{\mathcal{H}}^2 \leq \norm{\Sigma}_{\lambda, H_2}^2$.
  As to $\varphi^{T, u, p}$, due to $\p$ taking values in $[-1,1]^{n_{\text{p}}}$,
  $\norm{\varphi^{\lambda}_{i_q, i_r, I}(\tau)}_2^2 \le \norm{u(T-\tau_{k+1})}_{2}^2 e^{-\lambda \tau_{k+1}}$, for any
  $I \in I_k$,
  $\tau=(\tau_{k+1},\ldots,\tau_1) \in \Delta_{k+1}^T$, $k \ge 0$.
  Hence, by setting
  $\int_{\Delta_k^t} d\bt=1$ for $k=0$ and any $t > 0$, we obtain
  \begin{align*}
     &\norm{\varphi^{T, u, p}}_{\mathcal{H}}^2 = 
     \sum\limits_{i_q,i_r = 0}^{n_{\text{p}}}
     \sum\limits_{k = 0}^{\infty}\sum\limits_{I \in I_k}
     \int_{\Delta_{k+1}^T} 
     \norm{\varphi^{\lambda}_{i_q, i_r, I}(\tau)}_2^2
     \, d\bt  \\
     & \leq (n_{\text{p}} + 1)^2 \int_{0}^{T}
     \norm{u(T-t)}_2^2 \e^{-\lambda  t}
     \Bigg(\sum\limits_{k = 0}^{\infty}  
     \sum\limits_{I \in I_k}
     \int_{\Delta_k^t} \,d \tau \Bigg) dt 
\\
     & \leq (n_{\text{p}} + 1)^2 \int_0^T \norm{u(T-t)}_2^2 e^{t(n_{\text{p}}-\lambda)} dt  \\
     & \leq  (n_{\text{p}} + 1)^2 \norm{u}_{L^2([0,T], \mathbb{R}^{n_{\text{in}}})}^2 < (n_{\text{p}}+1)^2 L^2_{u} < +\infty. 
  \end{align*}
  The last inequality follows from the well-known upper bound
  $\int_{\Delta_k^{t}} \,d \tau  \le \frac{t^k}{k !}$ for iterated integrals
  \cite[Chapter 3.1]{isidori1985nonlinear}, leading to $\sum_{k=0}^{\infty}
  \sum\limits_{I \in I_k} \int_{\Delta_k^{t}} \,d \tau \, \le
  \sum_{k=0}^{\infty} \frac{n_{\text{p}}^k t^k}{k !}=e^{n_{\text{p}} t}$, as well as the choice of
  $\lambda \geq n_{\text{p}}$ and Assumption \ref{ass:tech}. Finally, by Lemma
  \ref{lemma:volterra}, $y_{\Sigma}(u, p)(T)
   = \langle \w^{T, \Sigma}, \varphi^{T, u, p} \rangle_{\mathcal{H}}$.
\end{altproof}}

Using \textcolor{blue}{Lemma \ref{lemma:main} }and  \cite[Lemma 26.10]{shalev2014understanding} we have
$R_{S}(\mathcal{F}) \leq \frac{c_{\mathcal{E}} L_{u}(n_{\text{p}} + 1)}{\sqrt{N}}$
yielding to $R_{S}(L_0(T)) \leq K_{\ell}R_{S}(\mathcal{F})
\leq \frac{c}{\sqrt{N}}$.

\textbf{Step 2: Bounding $B(T)$.}
By Assumption \ref{ass:tech}, we have
\begin{align*}
    |\ell(\ysh(u, p)(T), y(T))|
    &\leq 2K_{\ell}\max\{|\ysh(u, p)(T)|,|y(T)|\}\\
    &\leq 2K_{\ell}\max\{L_{u} c_{\mathcal{E}}, c_{y}\}  \leq c
\end{align*}
The last inequality follows from applying Lemma \ref{lemma:h2norm} and 
Assumption \ref{ass:tech} along with $n_{\text{p}} \geq 1$.

Finally, the proof of Theorem \ref{thm:main} follows from the bounds obtained in
Step 1. and Step 2. together with Theorem \ref{pac}.

\begin{color}{blue}
\section{Experiments}
\label{sec:experiments}
We considered a parameterized family  $\mathcal{E}$ of LPV systems
$\Sigma(\theta)=(A_i(\theta),B_i(\theta),C_i(\theta))_{i=0}^{n_{\text{p}}}$
where $n_{\text{p}}=1$, $\Theta \subseteq \mathbb{R}^3$, and 
for all $\theta=\begin{bmatrix} \theta_1 & \theta_2 & \theta_3 \end{bmatrix}^T \in \Theta$, 
\[  
\begin{split}
& A_0(\theta)=\begin{bmatrix} -\frac{1}{\theta_1} & 0  \\ 
1 &  \frac{-1}{\theta_1} 
\end{bmatrix},~ 
~ A_1(\theta)=\begin{bmatrix} 0 & \theta_2  \\ 
0 & 0 
\end{bmatrix},~   B_0(\theta)=\begin{bmatrix} \theta_3 \\ 0 \end{bmatrix}, \\
 & B_1(\theta)=\begin{bmatrix} 0 & 0 \end{bmatrix}^T, \quad  C_0(\theta)=\begin{bmatrix} 0 & 1 \end{bmatrix}, \quad  
C_1(\theta)=\begin{bmatrix} 0 & 0 \end{bmatrix} \\
& \Theta =\{ \theta \in \mathbb{R}^{3} \mid  
\|\Sigma(\theta)\|_{H_2,\lambda} < c_{\mathcal{E}} \}, ~  \lambda=1.2 \mbox{ and } c_{\mathcal{E}}=2.
\end{split}
\]
We considered the training data set $\mathbf{S}=\{u_i,p_i,y_i\}_{i=1}^{N}$,
and the validation data set $\mathbf{S}_{val}=\{u_i,p_i,y_i\}_{i=1}^{M}$,
$M=10^4$ where
$u_i,p_i$ are signals defined on $[0,T]=[0,0.45]$, $u_i,p_i$ are right continuous and they are constant on each interval $[kT_s,(k+1)T_s)$,
$k=0,\ldots,45$, $T_s=0.01$. Moreover, the values $p_i(kT_s)$ and $u_i(kT_s)$, $i=1,\ldots,N$, $k=1,\ldots, 45$ 
were sampled independently from the uniform distribution 
on $[0,3]$ and $[0,30]$ respectively and their means were substracted
to render them zero mean. 
The signal $y_i$ is the output of the system $\Sigma(\theta_{\star})$,
where $\theta_{\star}=\begin{bmatrix} 0.1 & -1.85 &-153.15 \end{bmatrix}^T$, to which we added a zero mean Gaussian white noise with variance $0.05$.
For simulating $\Sigma(\theta_{\star})$ we used Euler method with the discretization step $T_s=0.01$, which is the same as the sampling step $T_s$. The discrete-time LPV system corresponding  to discretizing 
$\Sigma(\theta)$ in time using Euler method and sampling rate $T_s$ admits the following input-output representation
$y_k=\phi_k a(\theta)$ such that
\begin{align*}
\phi_k& =\begin{bmatrix}  y_{k-1} & y_{k-2} & y_{k-2}p_{k-2} & u_{k-2} \end{bmatrix} \\
a(\theta)&=\begin{bmatrix} 2(1-\frac{T_s}{\theta_1}), & (1-\frac{T_s}{\theta_1})^2, & T_s^2\theta_2, & \theta_3 T_s^2
 \end{bmatrix}^T
\end{align*}
where $y_k,p_k,u_k$
are the output, scheduling signal, and input of the sampled time
version of $\Sigma(\theta)$ at sample time $k$, 
see \cite{arxiv} or the derivation  \cite[Section 4.2]{LPVtools}.
For every $N$, we use the input-output equation 
$y_k=\phi_k a(\theta)$ to estimate $a(\theta)$, using linear least squares, 
and we estimate $\theta$ using the estimates of $a(\theta)$, see \cite[Appendix D]{arxiv}
for details.

Numerical verification reveals that 
for each $N$, $\theta$ belongs to $\Theta$. 
We chose $\delta=0.05$ and we computed $R(\delta)$. We then computed 
$\mathcal{L}_N^{\mathbf{S}}(\Sigma(\theta))$, 
$\mathcal{L}(\Sigma(\theta))$,
using the norm as loss $\ell(y,y')=\|y-y'\|_2$, and
by approximating $\mathcal{L}(\Sigma(\theta))$
by the empirical prediction error $\mathcal{L}_M^{\mathbf{S}_{val}}(\Sigma(\theta))$ on the validation data set.
As Figure \ref{fig} shows,
$\mathcal{L}(\Sigma(\theta)) \leq  \mathcal{L}^{S}(\Sigma(\theta)) + \frac{R(\delta)}{\sqrt{N}}$ holds, 
and the resulting bound on the true loss 
$\mathcal{L}(\Sigma(\theta))$ is not vacuous: the bound for $N=10000$ is smaller
than the true loss $\mathcal{L}(\Sigma(\theta))$ for $N=200$.
\begin{figure}[t]
\includegraphics[width=6cm]{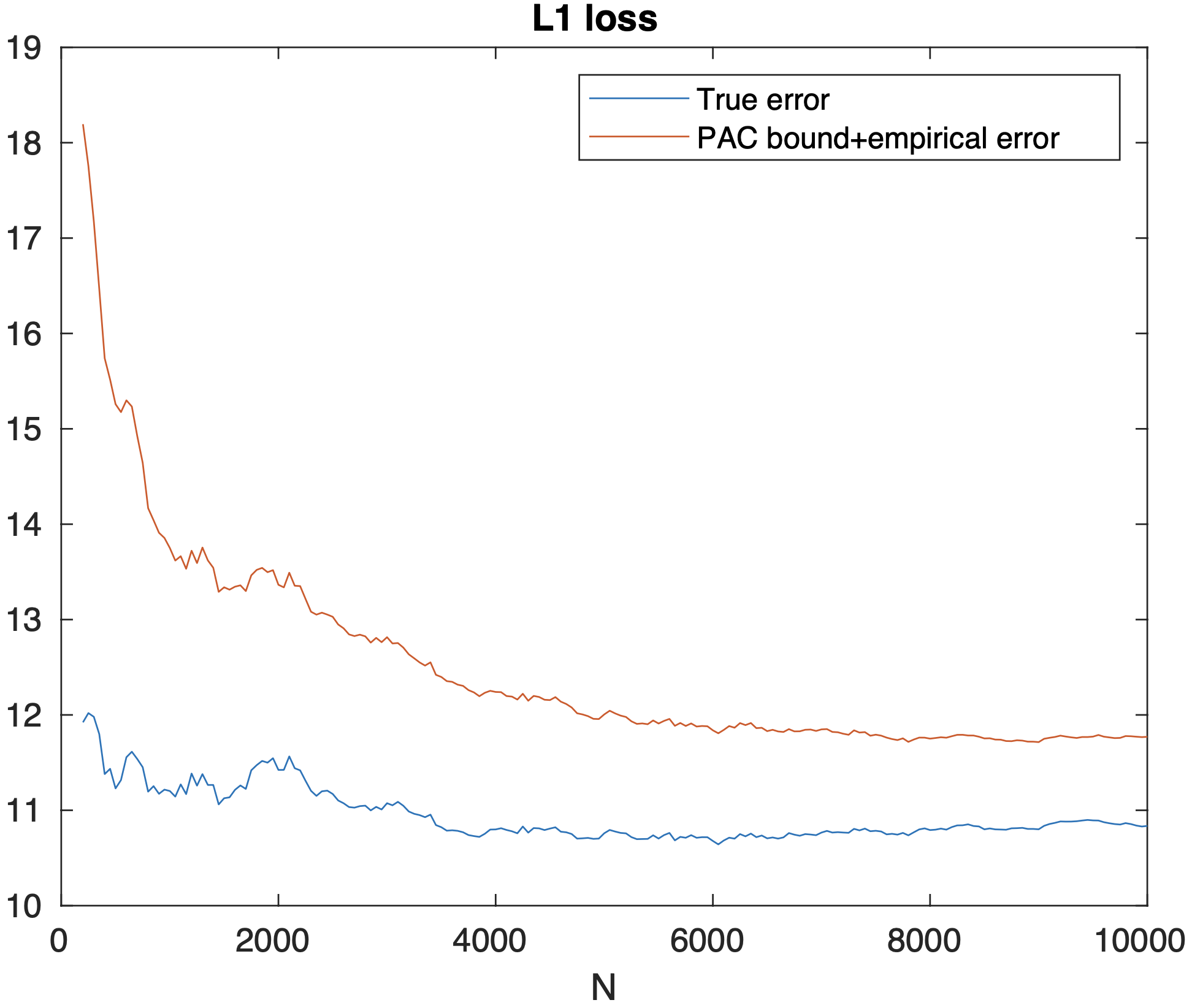}
\centering
\caption{\textcolor{blue}{Difference between the estimated and the true error. The estimated error is the sum of the empirical error measured on a finite set with $N$ elements and the PAC bound.  }} \label{fig}
\end{figure}
\end{color}

\section{Conclusion}
In this paper we examined LPV systems within the confines of statistical
learning theory and derived a PAC bound on the generalization error under
stability conditions. The central element of the proof is the application of
Volterra series expansion in order to upper bound the Rademacher complexity of
LPV systems. Further research is directed towards extending these methods to
more general models, possibly exploiting the powerful approximation properties
of LPV systems.

\bibliographystyle{ieeetr} 
\bibliography{references}



\section{Appendix A: Proof of Lemma \ref{lemma:volterra}}
\label{appA}

\begin{altproof}
    Consider the following bilinear system for all $i_q, i_r \in [n_{\text{p}}]$ for a
    fixed $t \in [0, T]$.
    \begin{equation}
    \label{eq:system3}
        \begin{split}
            &\dot{s}(\tau) = \Bigg (A_0 + \frac{\lambda}{2}I
            + \sum\limits_{i=1}^{n_{\text{p}}}p_i(\tau+T-t) A_{i} \Bigg) s(\tau),  \\
            &y^{s,t}(\tau) = C_{i_q} s(\tau), ~ s(0)
            = B_{i_r} u(T-t)e^{-\frac{\lambda}{2}t}.
        \end{split}
    \end{equation}
    From the  Volterra series representation
    \cite{isidori1985nonlinear} of bilinear systems we have 
    \begin{align*}
        y^{ s,t}_{i_q, i_r}(\tau) &=  
        \Big[w_{i_q,i_r,\emptyset}(\tau) + \\
        & \sum\limits_{k = 1}^{\infty} 
        \sum \limits_{\mathbf{I} \in I_k}
        \int_{\Delta_k^{\tau}}
        \w_{i_q,i_r,\mathbf{I}}^{\lambda,\tau}(\bt)
        p_{\mathbf{I}}(\bt) d\bt \Big] u(T-t)e^{- \frac{\lambda}{2} t}
    \end{align*}
    where $p_{I}$ is as in Section \ref{sec:ass} and
    $w_{i_q,i_r,I}^{\lambda,\tau}(\bt)=
    w_{i_q,i_r,I}^{\lambda}((\tau,\bt))$,
    for all $I \in I_k$ and $\bt \in \Delta_k^{\tau}$, $k  >0$. By
    \cite[Chapter 3.3.1.1]{toth2010modeling}
    $$
      \ysh(u, p)(T)\!\!=\!\!\int_{0}^{T}\!\!
       C(p(T))\Phi(T,T-t)B(\p(T-t))u(T-t)dt
    $$
    where $\Phi(s,s_0)$
    is the fundamental matrix of $\dot z(s)=A(p(s))z(s)$. Since the fundamental
    matrix $\Phi_{\lambda}(s,s_0)$ of $\dot z(s)=(A(p(s))+\frac{\lambda}{2}
    I)z(s)$ satisfies $\Phi_{\lambda}(s,s_0)=e^{\frac{\lambda}{2}(s-s_0)}
    \Phi(s,s_0)$, we have
    $s(\tau)=\Phi_{\lambda}(T-t+\tau,T-t)B_{i_r} u(T-t)e^{-\frac{\lambda}{2}t}
    = e^{\frac{\lambda}{2} (t-\tau)}\Phi(T-t+\tau,T-t)B_{i_r} u(T-t)$. Then
    from the definition of $C(p(T))$ and $B(p(T-t))$,
    $$\ysh(u, p)(T) = \int_{0}^{T} \sum\limits_{i_q, i_r = 0}^{n_{\text{p}}}
    p_{i_q}(T)p_{i_r}(T-t) y_{i_q, i_r}^{s,t}(t) \, dt.$$
    Applying the Volterra expansion above together with\\$\{ (t,\bt) \mid t
    \in [0,T], \bt \in \Delta_k^t\}=\Delta^{T}_{k+1}$ yields the result.
\end{altproof}

\section{Appendix B: proof of Lemma \ref{lemma:h2norm} and bounding the $H_2$ norms}
\label{appB}

\begin{altproof}
  If Assumption \ref{ass:tech} 
  holds, then $A_0^T Q + Q A_0 \prec -\lambda Q$, and hence $A_0+\lambda I$ is
  Hurwitz. Let $S= A_0^T Q + Q A_0 + \sum\limits_{i=1}^{n_{\text{p}}} A_i^T Q A_i +
  \sum\limits_{i=1}^{n_{\text{p}}} C_i^T C_i + C_0^T C_0+\lambda Q$. Then $S \prec 0$
  and hence $S=-VV^T$ for some $V$. Define $\tilde{C}=\begin{bmatrix} C_0^T &
  \ldots & C_{n_{\text{p}}}^T & V^T \end{bmatrix}^T$, $\tilde{A}=(A+0.5\lambda I)$ and
  $N_i=A_i$, $i \in [n_{\text{p}}]$. Then $\tilde{A}^T Q + Q \tilde{A}+\sum_{i=1}^{n_{\text{p}}}
  N_i^T Q N_i + \tilde{C}^T\tilde{C}=0$ and hence, by \cite[Theorem 6]{zl02}, for
  any choice of the matrix $G=\begin{bmatrix} G_1 & \ldots & G_{n_{\text{p}}}
  \end{bmatrix}$,
  the bilinear system 
  \begin{equation*}
      S(G) \left \{
      \begin{array}{rl}
      \dot z(t) &= \tilde{A}z(t)+\sum_{i=1}^{n_{\text{p}}}
       \left(N_i z(t)  p_i(t) + G_i p_i(t) \right)  \\ 
      \tilde{y}(t) &= \tilde{C}z(t), ~  z(0)=0,
      \end{array}\right.  
  \end{equation*}
  has a finite $H_2$ norm $\norm{S(G)}_{H_2}$ which satisfies
  $\norm{S(G)}_{H_2}^2=\mathrm{trace}(G^T Q G)$, and which is defined via Volterra
  kernels as follows. For each $i \in [n_{\text{p}}]$, let
  $g_{i,\emptyset}^G(t)=\tilde{C} e^{\tilde{A}t} G_i$ and for every $I
  \in I_k$, $k > 0$, let $g_{i, I}^G(t)
  =\tilde{C} e^{\tilde{A} t_{k+1}} N_{i_k} e^{\tilde{A} t_k} \cdots
  e^{\tilde{A}t_2} N_{i_1} e^{\tilde{A} t_1}G_{i}$,
  $t=(t_{k+1},\ldots,t_1) \in \mathbb{R}^k$, $t \in \mathbb{R}$. Then
  \begin{align*}
    \norm{S(G)}_{H_2}^2&= 
    \sum_{i=1}^{n_{\text{p}}} 
    \sum_{k=0}^{\infty}
    \sum_{I \in I_k}
    \int\limits_{[0,+\infty)^{k+1}}
    \norm{g_{i,I}^G(t)}_2^2dt.
  \end{align*}
  Let $G^{i,j}$ be the matrix of which the first column is the $j$th column of $B_i$
  and all the other elements are zero, $i \in [n_{\text{p}}]$, $j  \in [n_{\text{in}}]$.
  By choosing $G=G^{i,j}$, it
  follows that the $j$th row of $w_{i_q, i_r, I}^{\lambda}(\tau)$ is the
  $(i_q+1)$-th row of $g^{G^{i,j}}_{i_r,I}(t)$ where $\tau \in
  \Delta_{k+1}^{\infty}$ and
  $t=(\tau_{k+1}-\tau_{k},\ldots,\tau_2-\tau_1,\tau_1)$, $i_q,i_r \in
  [n_{\text{p}}]_0$, $I \in I_k$, $k\ge 0$.
  Hence, 
  $\sum_{j=1}^{n_{\text{in}}} 
  \norm{g_{i_r,I}^{G^{i_r,j}}(\tau)}_2^2
  \ge \sum_{i_q=0}^{n_{\text{p}}} 
  \norm{w_{i_q, i_r,
  I}^{\lambda}(\tau)}_2^2$, and
  by applying a change of variables in the iterated integrals, 
  \begin{align*}
    & \sum_{i_q,i_r=0}^{n_{\text{p}}}
    \sum_{k=0}^{\infty} 
    \sum_{I \in I_k}
    \int_{\Delta_{k+1}^{t}} 
    \norm{w^{\lambda}_{i_q, i_r, I}(\tau)}^2_2 d\tau 
    \le \\
    & \sum_{i_r=0}^{n_{\text{p}}} \sum_{j=1}^{n_{\text{in}}} \norm{S(G^{i_r,j})}_{H_2}^2
    = \sum_{i_r=0}^{n_{\text{p}}} \mathrm{trace}(B_{i_r}^T Q B_{i_r}).
  \end{align*}
  In the last step we used the fact that $\sum_{j=1}^{n_{\text{in}}}
  \mathrm{trace}((G^{i_r,j})^T Q G^{i_r,j})=\mathrm{trace}(B_{i_r}^T Q B_{i_r})$

Finally, from Lemma
  \ref{lemma:main} it follows that $y_{\Sigma}(u,p)(T)= \langle \w^{T, \Sigma},
  \varphi^{T, u, p} \rangle_{\mathcal{H}}$ for a suitable Hilbert space. From
  the definition of $\w^{T,\Sigma}$ and  $\varphi^{T, u, p}$ and the proof of
  Lemma \ref{lemma:main} it follows that $\|\w^{T,\Sigma}\|_{\mathcal{H}} \le
  \|\Sigma\|_{\lambda,H_2}$, $\|\varphi^{T,u, p}\|_{\mathcal{H}} \le (n_{\text{p}}+1)
  \|u\|_{L^2([0,T],\mathbb{R}^{n_{\text{in}}})}$ and hence by the Cauchy-Schwartz
  inequality the result follows.
\end{altproof}


\textbf{Bounding the $H_2$ norms.}
Below we present a sufficient
condition for computing an upper bound 
on the $H_2$ norms of models from $\mathcal{E}$. 
\begin{assumption}
    \label{ass:extra}
    There exists a positive real number $\gamma > 0$ and $\Gamma > 0$ such that
    $\gamma \geq \Gamma+n_{\text{p}}$ and for every $\Sigma \in \mathcal{E}$ of the form
    \eqref{system:2}, $\Gamma \ge \sup_{i=1}^{n_{\text{p}}} \norm{A_i}^2_{2}n_{\text{p}}$ and
    $\norm{\e^{A_0 t}}_2 \leq \e^{-\frac{\gamma}{2} t}$.
\end{assumption}

\begin{lemma}
\label{lemma:sigma}
 If Assumption \ref{ass:extra} holds then \eqref{ass:stability} holds  with any $n_{\text{p}}
 \le \lambda \le \gamma-\Gamma$ and $Q = I$, and 
 \begin{align*}
     c_{\mathcal{E}} \le 
         (n_{\text{p}}+1)^2 K^2_C K^2_B
          \Big(\frac{1}{\lambda}  + \frac{1}{\gamma - \lambda + \Gamma } \Big)
 \end{align*}
where $K_B=\sup_{i=1}^{n_{\text{p}}} \norm{B_i}_2$, $K_C=\sup_{i=1}^{n_{\text{p}}} \norm{C_i}_2$.
\end{lemma}

The proof follows from a simple calculation followed by
applying Lemma \ref{lemma:h2norm}.

\section{Appendix C: multi output}

Let use assume that the loss functions satisfies the following condition:
$\ell(y,\tilde{y})=\sum_{j=1}^{p} \ell_i(y^j,\tilde{y}^j)$, 
where $y^j,\tilde{y}^j$ denote the $j$th component of the vectors
$y^j, \tilde{y}^j \in \mathbb{R}^{p}$ respectively and $\ell_j$ are
Lipschitz functions with Lipschitz constant $K_{\ell}$.
For instance, if $\ell(y,\tilde{y})=\|y-y^{'}\|_1$ is the $\ell_1$ loss,
or $\ell(y,y^{'})=\|y-y^{'}\|_2^2$ is the classical quadratic
loss and $y,y^{'}$ are bounded, then this assumption is satisfied.
For all $\Sigma \in \mathcal{E}$, let 
$\Sigma_j$ be the LPV system which arises from $\Sigma$ by 
considering only the $j$th output, and let $\mathcal{E}_j$ be the class of LPV
systems formed by all  $\Sigma_j$, $\Sigma \in \mathcal{E}_j$.  For every $j=1,\ldots, n_{\text{out}}$, consider the data set 
$\mathbf{S}_j = \{(u_i, p_i, y_i^j) \}_{1 \leq i \leq N}$,
which is obtained from the data set $\mathbf{S}$
by taking only the $j$th component of the 
true outputs $\{y_i\}_{i=1}^{n}$. 
Then it follows that $\mathcal{L}(\Sigma)=\sum_{j=1}^{n_{\text{out}}} \mathcal{L}(\Sigma_j)$,
where $\mathcal{L}(\Sigma_j)=\mathbb{E}_{(u, p, y) \sim \mathcal{D}}[\ell_j(
  y_{\Sigma_j}(u,p)(T),y^j(T))]$
and $\mathcal{L}_{N}^{\mathbf{S}}(\Sigma)=\sum_{j=1}^{n_{\text{out}}} \mathcal{L}_N^{\mathbf{S}_j}(\Sigma_j)$. By applying the main theorem of the paper to
$\mathbf{S}_j$ and $\mathcal{E}_j$, $j=1,\ldots,n_{\text{out}}$, it follows that
\begin{align*}
        \mathbb{P}_{\mathbf{S} \sim
        \mathcal{D}^N}\Bigg(\forall \Sigma_j \in \mathcal{E}_j:
        \mathcal{L}(\Sigma_j) - \mathcal{L}^{\mathbf{S}_j}_{N}(\Sigma_j)
        \leq \frac{R_j\delta)}{\sqrt{N}} \Bigg) \geq 1 - \delta.
\end{align*}
for all $j=1,\ldots, n_{\text{out}}$. Then by using the union bound it follows that
\begin{align*}
        &\mathbb{P}_{\mathbf{S} \sim
        \mathcal{D}^N}\Bigg(\forall j=1,\ldots,n_{\text{out}}, ~  \forall \Sigma_j \in \mathcal{E}_j:
        \mathcal{L}(\Sigma_j) - \mathcal{L}^{\mathbf{S}_j}_{N}(\Sigma_j) \\
        &\leq \frac{\max_{1 \le j \le n_{\text{out}}} R_j(\delta)}{\sqrt{N}} \Bigg) \geq 1 - n_{\text{out}}\delta.
\end{align*}
and by using $\mathcal{L}(\Sigma)=\sum_{j=1}^{n_\text{out}} \mathcal{L}(\Sigma_j)$
and $\mathcal{L}_{N}^{\mathbf{S}}(\Sigma)=\sum_{j=1}^{n_{\text{out}}} \mathcal{L}_N^{\mathbf{S}_j}(\Sigma_j)$ it then follows that 
\begin{align*}
        &\mathbb{P}_{\mathbf{S} \sim
        \mathcal{D}^N}\Bigg(\forall \Sigma \in \mathcal{E}:
        \mathcal{L}(\Sigma) - \mathcal{L}^{\mathbf{S}}_{N}(\Sigma) \\
        &\leq \frac{\max_{1 \le j \le n_{\text{out}}} R_j(\delta)}{\sqrt{N}} \Bigg) \geq 1 - n_{\text{out}}\delta.
\end{align*}

\section{Appendix D: Computing $\theta$ in numerical example}

We compute the estimate $\hat{\theta}$  of $\theta$  so that
$a(\hat{\theta})$, 
is the least squares solution of
$y_{i,k}=\begin{bmatrix}  y_{i,k-1} & y_{i,k-2} &  y_{i,k-2}p_{i,k-2} &  u_{i,k-2} \end{bmatrix} a(\theta)$,
$i=1,\ldots,N$, $k=45$, where 
where the function function $a(\theta)$ is 
\[ 
  a(\theta)=\begin{bmatrix} 2(1-\frac{T_s}{\theta_1}), & (1-\frac{T_s}{\theta_1})^2, & T_s^2\theta_2, & \theta_3 T_s^2
 \end{bmatrix}^T,
\]
$y_{i,l}=y_i(lT_s)$, $u_{i,l}=u_i(lT_s)$, and 
$p_{i,l}=p_i(lT_s)$ for all $l \in \mathbb{N}$.

\section{Appendix E: difference between minimal prediction error models and best possible models}
Consider the minimal prediction error model
$\hat{\Sigma}=\mathrm{argmin}_{\Sigma \in \mathcal{E}}\mathcal{L}^S_N(\Sigma)$  and the best possible model:  $\Sigma_{\star}=\mathrm{argmin}_{\Sigma \in \mathcal{E}} \mathcal{L}(\Sigma)$, see \cite{arxiv}.

\begin{corollary}
\label{main:col1}
With the notation of Theorem \ref{thm:main} and
with $R(\delta) := c \left (2 + 5\sqrt{2\log(8/\delta)}\right )$ , for any $\delta \in [0,\delta]$
\begin{align*}
        \mathbb{P}_{\mathbf{S} \sim
        \mathcal{D}^N}\Bigg(\mathcal{L}(\hat{\Sigma}) \le \mathcal{L}(\Sigma_{\star}) + \frac{R_2(\delta)}{\sqrt{N}} \Bigg) \geq 1 - \delta.%
\end{align*}
\end{corollary}
That is, we can determine the minimal number of
data points $N_{min}=(\frac{R_2(\delta)}{\epsilon})^2$, such that the difference between the 
performance of the minimal
prediction error model and that of the best possible model is below the desired  threshold $\epsilon$.  
\end{document}